# Comparative uncertainty, belief functions and accepted beliefs

**Didier Dubois**  **Hélène Fargier**  **Henri Prade**
Institut de Recherche en Informatique de Toulouse
Université Paul Sabatier – 118 route de Narbonne
31062 Toulouse Cedex 4 – France
Email: {dubois, fargier, prade}@irit.fr

## Abstract

This paper relates comparative belief structures and a general view of belief management in the setting of deductively closed logical representations of accepted beliefs. We show that the range of compatibility between the classical deductive closure and uncertain reasoning covers precisely the nonmonotonic 'preferential' inference system of Kraus, Lehmann and Magidor and nothing else. In terms of uncertain reasoning any possibility or necessity measure gives birth to a structure of accepted beliefs. The classes of probability functions and of Shafer's belief functions which yield belief sets prove to be very special ones.

## 1. INTRODUCTION

There is an old controversy in the framework of Artificial Intelligence between probabilistic and other numerical representations of uncertainty on the one hand, and the symbolic setting of logical reasoning methods on the other hand. The emergence and relative success of some numerical reasoning formalisms based on probability (especially Bayesian networks) or fuzzy sets has led AI to accept numerical representations as complementary to symbolic ones. However the basic issue underlying the controversy, that is to say, whether or not the two forms of reasoning (logical and numerical) are at all compatible at the formal level, has not been widely addressed. Namely if the beliefs of an agent are represented simultaneously by a measure of confidence (such as a probability function for instance) on a set of states and by a logical knowledge base that accounts for the propositions sanctioned by this measure of confidence, two questions are worth investigating:
   • to what extent are inferences drawn from the measure of confidence and inferences drawn from the logical knowledge base coherent?
   • to what extent can the revisions of the measure of confidence (via conditioning) and of the knowledge base, upon arrival of an input information, be coherent as well?
These questions have been considered in the past only indirectly and with respect to particular uncertainty theories, basically probability theory (Kyburg 1988). This paper addresses the problem by trying to relate a general view of belief revision in the setting of deductively closed logical representations (belief sets of Gärdenfors, 1988), and comparative belief structures, that is relations comparing events, in terms of relative likelihood, certainty, plausibility and the like. Such relations can be induced by *numerical* representations of belief such as probability, possibility (Lewis 1973; Dubois 1986), Shafer's belief functions (Wong et al. 1991). The single assumption of deductive closure for belief sets severely restricts the type of comparative belief structure that can be envisaged for reasoning under uncertainty. Actually, we show that the range of compatibility between the classical deductive closure and uncertain reasoning precisely covers the kind of nonmonotonic inference system called preferential by Kraus, Lehmann and Magidor (1990), and nothing else.

Our work is closely related to findings of Halpern (1996) who investigates how to lift a partial order on states to a partial order on events in the style of possibility theory and of Friedman and Halpern (1996) that describe the semantics of nonmonotonic inference in terms of confidence relations. The present paper makes a step further: it considers a larger class of confidence relations and extends Friedman and Halpern's results about the representation of partial preorders that are in accordance with preferential entailment. Moreover, while (Friedman and Halpern 1996) mainly focuses on equipping preferential inference with a semantics in terms of uncertainty relations, we are also interested in checking the consistency between the notion of logical closure and the various numerical theories of uncertainty: probability measures, possibility and necessity measures (Zadeh 1978), belief and plausibility functions (Shafer 1976). Thus, the paper should be also viewed as a continuation of a previous paper by two of the authors (Dubois, Prade 1995b).

Section 2 gives the background on confidence relations and belief revision. Section 3 introduces the general framework of partial acceptance preorders, which are partial preorders compatible with logical consequence and conditioning; examples and discussions are provided; links with system P are presented in the last part of Section 3. Section 4 studies characteristic properties and consequences of additivity in this framework, and a representation theorem of any acceptance preorder by a family of complete acceptance preorders is given. Section 5 focuses on complete acceptance preorders, and more precisely on those that follow from well-known numerical confidence measures.

## 2 BACKGROUND

### 2.1 CONFIDENCE RELATIONS

Usually, in uncertainty theories, an agent builds a comparative belief structure, as a relation $\geq$ among *events* A, B, C, i.e., subsets of a set of states S (in the following, we assume that S is finite). $A \geq B$ means that the agent's confidence in A is at least as great as his confidence in B. From this confidence relation, associated relations are defined in a classical way:
- $A > B$ iff $(A \geq B)$ and not $(B \geq A)$ - *strict preference;*
- $A \not\Vert B$ iff neither $(A \geq B)$ nor $(A \geq B)$ - *incomparability;*
- $A \equiv B$ iff $(A \geq B)$ and $(B \geq A)$ - *equivalence.*

Notice that, in order to tell incomparability from equivalence, we do not suppose that $\geq$ is complete (this generalizes usual confidence measures, that define a *complete* preorder, i.e. where either $A \geq B$ or $B \geq A$ since mapping to a numerical scale). A confidence relation is naturally transitive but the incomparability relation it induces is not necessarily so. Moreover if $A \subseteq B$, then A and B are comparable and confidence in A cannot be strictly greater than the one in B: this forbids situations where a proposition would be more believed than another it semantically entails. In summary the main assumption concerning a confidence relation are:

*Transitivity (T):* $A \geq B$ and $B \geq C$ imply $A \geq C$

*Monotony with respect to inclusion (MI) :*
$$A \subseteq B \Rightarrow B \geq A.$$

MI is a weak form of the orderly axiom of Halpern (1996):

*Axiom O: if $A \subseteq A'$ and $B' \subseteq B$ then $A > B \Rightarrow A' > B'$*

The following properties follow from T and MI :
- Reflexivity of $\geq$ and thus of $\equiv$;
- Transitivity of $>$ and $\equiv$;
- Irreflexivity of $>$ (IR) : $A > A$ does not hold ($A > A$ makes no sense);
- Compatibility with logical entailment for both $\geq$ and $>$:
  if $A \subseteq A'$ and $B' \subseteq B$ then $A \geq B \Rightarrow A' \geq B'$.
if $A \subseteq A'$ and $B' \subseteq B$ then $A > B \Rightarrow A' > B'$ (Axiom O).

Some confidence relations are usually specified by defining a stability property on the relation $A \geq B$ with respect to adding new elements to sets A and B or deleting common elements (Fishburn 1986; Dubois 1986) :

**Definition 1** *(additivity). Let A, B and C be any three events such that $A \cap (B \cup C) = \emptyset$. A relation $\geq$ is additive iff $A \cup B \geq A \cup C \Leftrightarrow B \geq C$.*

Denoting $\geq^T$ the dual of $\geq$, such that $A \geq^T B$ iff $\bar{B} \geq \bar{A}$, it can be checked that additive relations are self dual, i.e., additive confidence relations are such $A \geq B$ iff $\bar{B} \geq \bar{A}$. For instance, comparative probabilities (Fishburn 1986) are additive confidence relations. There are well-known confidence relations that are not additive and not even self-dual, e.g., those based on Shafer's belief or plausibility functions. For instance, consider the basic probability assignment m with two disjoint focal elements E and F such that $m(E) < m(F)$, and two events A and B such as $E \subseteq A, A \cap B = \emptyset, F \cap A = \emptyset, F \cap B \neq \emptyset$ and $F \cap \bar{B} \neq \emptyset$; then $Bel(A) = m(E) > Bel(B)=0$ while $Pl(A) = m(E) < Pl(B) = m(F)$, in spite of the fact that Bel and Pl are dual. This would suggest, when $\geq$ is not self-dual, to study both the relation defined by $A \geq B$ and the one defined by $\bar{B} \geq \bar{A}$.

Besides, for confidence relations which are not self-dual, it is possible to add two further requirements in order to distinguish respectively between the two kinds of relations generated by duality, namely
- not($A \equiv \emptyset$ and $\bar{A} \equiv \emptyset$) for possibility-like relations
- not($A \equiv S$ and $\bar{A} \equiv S$) for certainty-like relations.

### 2.2 ACCEPTED BELIEFS AND BELIEF REVISION

The classical logic approach to belief representation considers a set **B** of propositions $\phi, \psi,...$ as a set of explicit beliefs of an agent, from which it is possible to infer implicit beliefs. The set of consequences of **B** is the set of "rationally" accepted beliefs also called a belief set **K** (Gärdenfors 1988). In order to relate belief sets and confidence relations, we identify events to sets $[\phi]$ of models of propositions $\phi$, and we must decide what it means for a proposition to be accepted by a confidence relation. The weakest and most natural way is to say that $\phi$ is an accepted belief if and only if the agent is more confident in $\phi$ than in its negation: $[\phi] > [\neg\phi]$ ( or equivalently $[\phi] > [\bar{\phi}]$ ). In this paper we shall use the event-like notation: $A > \bar{A}$ when A is accepted, which is also equivalent to $A >^T \bar{A}$.

Since a belief set is deductively closed, the compatibility between the confidence relation and the belief set requires that any consequence of an accepted belief is an accepted belief and that the conjunction of two accepted beliefs is an accepted belief (Dubois, Prade 1995b):

- Consequence stability (CS):
    if $A \subseteq B$ and $A > \bar{A}$ then $B > \bar{B}$
- AND rule: if $A > \bar{A}$ and $B > \bar{B}$ then $A \cap B > \bar{A} \cup \bar{B}$.

A belief is tentatively accepted and may be questioned by the arrival of new information. Hence the compatibility between belief sets and confidence relations must take into account the dynamics of belief. When an input information arrives as a proposition $\psi$ that must be true, the belief set **K** is revised into **K***$\psi$, that is, a belief set that contains $\psi$. Rationality axioms which specify how a revision operator should behave, especially when the initial belief set is contradictory with $\psi$, have been proposed (Gärdenfors 1988). Here we shall assume nothing but the success postulate ($\psi \in \mathbf{K}*\psi$) and the deductive closure of **K***$\psi$.

Revising a confidence relation by an input C = [ψ] comes down to restricting the relation to subsets of C. This is called conditioning. There is more confidence in A than in B, conditioned on C if and only if $A \cap C > B \cap C$. The set of accepted beliefs when C holds is thus $A_>(C) = \{A: A \cap C > \overline{A} \cap C\}$[1]. The close connection between belief revision theory and conditioning is noticeable since, using Grove's dual $>_G$ of an epistemic entrenchment (Gärdenfors 1988) and a belief set K, the set of accepted events corresponding to the revised belief set K*ψ by an input ψ is precisely $A_{>_G}(C) = \{A: A \cap C >_G \overline{A} \cap C\}$ (with C = [ψ] ). Note that in terms of the epistemic entrenchment itself ($A >_E C$ iff $\overline{C} >_G \overline{A}$) we also have $A_{>_G}(C) = \{A: \overline{A} \cup C >_E \overline{A} \cup \overline{C}\}$. More generally, the latter (with any confidence relation $>^T$ instead of $>_E$) could be used as an equivalent definition of $A_>(C)$. Mind that $A_>(C)$ is not equal to $A_{>^T}(C)$ in general. Compatibility with the revision of belief sets leads thus to require that $A_>(C)$ is deductively closed, as K*ψ is. It leads to require conditional versions of CS and the AND rule:

- Conditional consequence stability (CCS):
if $A \subseteq B$ and $C \cap A > C \cap \overline{A}$ then $C \cap B > C \cap \overline{B}$;

- Conditional AND rule (CAND):
if $C \cap A > C \cap \overline{A}$ and $C \cap B > C \cap \overline{B}$
then $C \cap A \cap B > C \cap (\overline{A} \cup \overline{B})$.

CCS and CAND give the CS and AND rules when C = S.

In this paper, we try to characterise the common ground between confidence relations and revisable belief sets, i.e., confidence relation satisfying T, MI and CAND (CCS is implied by MI). These kinds of orders, that may be partial, will be called partial acceptance preorders so as to stress their compatibility with closed sets of accepted beliefs.

## 3 PARTIAL ACCEPTANCE PREORDER

Let S be a set of states, $2^S$ a set of events, and ≥ be a partial order on $2^S$ that satisfies T and MI. It is easy to see that CCS and CAND can be written as follows (A, B, C, D being disjoint events):

- CCS:           $A > B \cup C \Rightarrow A \cup C > B$;
- CAND: $A \cup B > C \cup D$ and $A \cup C > B \cup D$
                               $\Rightarrow A > B \cup C \cup D$.

---

[1] This is a natural definition of revision on the basis of a confidence relation. Indeed, in numerical uncertainty theories, conditioning has been studied axiomatically by Cox (1961) using the following equation: g(A|C) is a solution to the equality g(A∩C) = g(A|C)*g(C) where * is continuous and strictly monotonic in both places. Then the only possible solution is * = product. Then g(A|C) > g($\overline{A}$|C) if and only if g(A∩C) > g($\overline{A}$∩C).The equivalence holds when g is a probability, a *plausibility* function or a *possibility* measure, using Dempster rule of conditionning or its qualitative counterpart in possibility theory setting, which remains in the spirit of Cox's above conditioning equation. However, the equivalence does not hold for the dual conditional measures of the form $g^T(A|B) = 1 - g(\overline{A}|B)$.

In fact, as soon as T and MI are assumed, it is enough to add the following axiom introduced in (Dubois, Prade 1995b; Friedman and Halpern 1996):

*Acceptance Axiom (Ac):* ∀A,B,C three disjoint events,
$A \cup B > C$ and $A \cup C > B \Rightarrow A > B \cup C$.

The CCS property is recovered from axiom MI and the CAND property is equivalent to Ac in presence of axiom MI. Thus, we will define partial acceptance preorders by:

**Definition 2:** *a relation ≥ on a set of events is an acceptance preorder iff it satisfies T, MI and Ac. The relation > induced by an acceptance preorder will be called a partial acceptance order.*

Note that the dual of an acceptance preorder is generally not an acceptance preorder, except of course for self-dual relations. Partial acceptance orders are characterized by:

**Proposition 1** : *Any partial acceptance order is irreflexive, transitive and satisfies O and Ac. Reciprocally, if ≻ is irreflexive, transitive and satisfies O and Ac then it is a partial acceptance order induced by the relation ≥ defined by $A \geq B$ iff $A \succ B$ or $B \subseteq A$. Relation ≥ is an acceptance preorder such that ($A \equiv B$ iff A=B).*

**Sketch of proof:** The strict part of an acceptance order is obviously irreflexive, transitive and satisfies O. It satisfies Ac by definition. Suppose that ≻ is an irreflexive and transitive that satisfies O and Ac. The relation ≥ defined in Proposition 1 satisfies MI by definition and clearly induces ≻. It obviously satisfies Ac. It is transitive since:
- if A ≻ B and B ≻ C, A ≻ C and thus A ≥ C
- if A ≻ B and C⊆B, by O, A ≻ C and thus A ≥ C
- if B⊆A and B ≻ C, by O, A ≻ C and thus A ≥ C
- if B⊆A and C⊆B, C⊆A and thus A ≥ C.

Moreover, A=B implies obviously A ≡ B. A ≡ B implies (A ≻ B or B⊆A) and (B ≻ A or A⊆B). Since A⊆B implies that not(A ≻ B), we have: B⊆A and A⊆B.

Moreover, any acceptance preorder satisfies the characteristic property of acceptance functions (Dubois, Prade 1995b):

**Proposition 2:** $\exists K \subseteq S, K \neq \emptyset$, *such that :*
$\forall A \subseteq S, A > \overline{A} \Leftrightarrow K \subseteq A$.

The set K is nothing but the intersection of all A's such that $A > \overline{A}$ and is the most specific belief induced by the partial acceptance order. We can also prove a new and more general property, that includes the dynamics of partial acceptance orders and which thus holds for context-tolerant acceptance functions:

**Proposition 3** $\forall B \subseteq S$, *if* $B > \emptyset$, *then* $\exists K_B \subseteq S$ *such that* $K_B \neq \emptyset$ *and* $\forall A \subseteq S, A \cap B > \overline{A} \cap B \Leftrightarrow K_B \subseteq A$

$K_B$ is actually the intersection of all the A such that $A \cap B > \overline{A} \cap B$. Proposition 2 is retrieved when B = S.

### 3.1 EXAMPLES

Well known examples of acceptance preorders are necessity and possibility orders (Dubois, 1986; Dubois, Prade 1995b). Let us recall that a necessity measure N is such that: $N(A) \in [0,1]$ and $N(A \cap B) = \min(N(A), N(B))$. The dual set function $\Pi(A) = 1 - N(\bar{A})$ such that $\Pi(A \cup B) = \max(\Pi(A), \Pi(B))$ is a possibility measure. More precisely, for dual pair(N, $\Pi$) of set functions, $\geq_N$ and $\geq_\Pi$ which are the dual necessity and possibility orders induced by N and $\Pi$ respectively are acceptance preorders.

Necessity orders are described by Gärdenfors (1988) under the name epistemic entrenchment. Indeed, any rational revision is driven by a complete preorder on events whose only numerical representatives are the necessity measures N of possibility theory. The set of accepted beliefs $A_\Pi = \{A, A >_N \bar{A}\} = \{A, A >_\Pi \bar{A}\}$ is deductively closed and is a semantic description of a belief set. Moreover the conditioned set $A_\Pi(C) = \{A, A \cap C >_\Pi \bar{A} \cap C\}$ is also deductively closed and satisfies all postulates of rational revision (Dubois, Prade 1992). Hence the strict part of a possibility order is a partial acceptance order.

Since the Gärdenfors revision theory underlies a particular confidence relation, it naturally suggests that any confidence relation might in turn lead to a revision theory for belief sets. However a strict order on events induced by a probability measure is generally *not* an acceptance order as it violates the AND rule. This has been noticed a long time ago by Kyburg (1961) and is the basis of the lottery paradox (one can believe that any given player in a one-winner game will lose, all the more so as players are numerous, but one cannot believe that all of them will lose).

Possibility orders and necessity orders are acceptance preorders, where *any* pair of events is comparable. Requiring the full comparability of events (i.e., requiring completeness) makes sense if the confidence order stems from a numerical set function, as studied in Section 5. But the incomparability is often not transitive. For instance, let $\Pi$ be a possibility function and consider the partial order $>_{\Pi sup}$ introduced in (Dubois et al 1997):

$A >_{\Pi sup} B$    iff $\Pi(A \cap \bar{B}) > \Pi(\bar{A} \cap B)$
            iff $N(A \cup \bar{B}) > N(B \cup \bar{A})$;
$A \geq_{\Pi sup} B$    iff $A >_{\Pi sup} B$ or $B \subseteq A$.

$\geq_{\Pi sup}$ is a partial acceptance preorder and $>_{\Pi sup}$ is a self-dual partial acceptance order that refines $>_\Pi$ ($A >_\Pi B$ implies $A >_{\Pi sup} B$) by comparing only the non-common part of events (if A and B are disjoint, $A >_{\Pi sup} B$ iff $A >_\Pi B$, and then $A_\Pi(C) = A_{\Pi sup}(C)$). But $\not\geq_{\Pi sup}$ is generally not transitive: $\Pi(A \cap \bar{B}) = \Pi(\bar{A} \cap B)$ and $\Pi(B \cap \bar{C}) = \Pi(\bar{B} \cap C)$ do not imply $\Pi(A \cap \bar{C}) = \Pi(\bar{A} \cap C)$.

Other examples of partial acceptance orders are orderly "qualitative" relations where a relation $>$ on $2^S$ is "qualitative" in the sense of Halpern iff:

• $\forall A, B, C \; A \cup B > C$ and $A \cup C > B \Rightarrow A > B \cup C$ (Qual)
• $>$ satisfies the orderly axiom O.

Halpern(1996) never supposes anything about reflexivity or irreflexivity. Anyway, Qual is much stronger than Ac since it applies to *any* A,B,C (not necessarily disjoint). Partial acceptance preorders encompass *reflexive* orderly qualitative relations and partial acceptance orders encompass irreflexive orderly relations. Indeed, any reflexive orderly qualitative relation is a an acceptance preorder and any irreflexive qualitative orderly relation is a partial acceptance order. But there are acceptance preorders[2] as well as partial acceptance orders[3] that are not "qualitative" in the sense of Halpern.

The framework of acceptance (pre)-orders thus takes a step further beyond orderly qualitative preference relations, be they reflexive or irreflexive. It does not encompass the case of relations that are neither reflexive, nor irreflexive - but can they be understood as confidence relations when $A > A$ for some A ?

### 3.2 RELATIONS WITH SYSTEM P

This framework is mainly the one proposed by Friedman and Halpern (1996), *except that* we do not need to assume their axiom A3 (that requires that $A \equiv B \equiv \emptyset$ implies $A \cup B \equiv \emptyset$). Indeed this axiom does not agree with belief functions or necessity measures for instance. Anyway, it remains true that any partial acceptance order satisfies the main axioms of system P and that, reciprocally, the order on disjoint events induced by a conditional knowledge base is an acceptance one.

More precisely, recall that several semantics of nonmonotonic logic lead to a strict order between *disjoint* events induced by a conditional knowledge base satisfying the postulates of preferential inference of (Kraus, Lehmann and Magidor 1990). Let $\phi \vdash \psi$ be a conditional assertion, stating that if $\phi$ holds then generally $\psi$ holds too. A partial order $>$ on disjoint events is induced by a set of conditional assertions by interpreting $\phi \vdash \psi$ as the statement that the conjunction $\phi \wedge \psi$ is more plausible than $\phi \wedge \neg \psi$:

$$\phi \vdash \psi \Leftrightarrow [\phi] \cap [\psi] > [\phi] \cap [\neg \psi] \quad (*)$$

In this context, well-known properties of nonmonotonic preferential inference can be written as properties of the confidence order $>$:

Or: $A \cap C > A \cap \bar{C}, B \cap C > B \cap \bar{C}$
$\Rightarrow (A \cup B) \cap C > (A \cup B) \cap \bar{C}$
Rw: $B \subseteq C, A \cap B > A \cap \bar{B} \Rightarrow A \cap C > A \cap \bar{C}$
Cm: $A \cap B > A \cap \bar{B}$ and $A \cap C > A \cap \bar{C}$
$\Rightarrow A \cap B \cap C > A \cap B \cap \bar{C}$
Cut: $A \cap B > A \cap \bar{B}$ and $A \cap B \cap C > A \cap B \cap \bar{C}$
$\Rightarrow A \cap C > A \cap \bar{C}$.

Kraus, Lehmann and Magidor (1990) also assume the reflexivity axiom $A \vdash A$, which is hard to accept in the present framework since it means $A > \emptyset$ for all A (and thus $\emptyset > \emptyset$!): this violates our antireflexivity requirement for $>$. Even the Restricted Reflexivity (RR) axiom (If $A \neq \emptyset, A > \emptyset$) is questionable in our context, since we aim at encompassing uncertainty theories: axiom RR indeed

---

[2] Consider for instance the reflexive relation $\geq$ on $S = \{s1, s2\}$ satisfying MI, whose strict part is $\{s1, s2\} > \{s1\} > \{s2\}$. Relation $\geq$ is an acceptance preorder that does not satisfy (QUAL). Indeed : $\{s1\} \cup \{s1\} \geq \{s2\}$ and $\{s1\} \cup \{s2\} \geq \{s1\}$ but $\{s1\} \geq \{s1\} \cup \{s2\}$ does not hold.

[3] For instance: $S = \{s1, s2\}, \{s1, s2\} > \{s1\} > \{s2\} > \emptyset$ satisfies AND, O, IR, T but not Qual.

enforces any non-contradictory event to be somewhat plausible. However, in the context of uncertainty theories, one may often wish to express that A is impossible even if A ≠ Ø (for instance, null sets in probability theory !). Anyway, this is easily remedied by introducing the following Consistency Preservation axiom, that is an obvious consequence of MI:

CP:       Whatever A, Ø > A does not hold.

It is easy to show that partial acceptance orders are induced by conditional assertions in the above sense and that it reciprocally turns out that the entailment relation induced by a partial acceptance order is not more general than preferential inference, but for the reflexivity A ⊢ A:

**Proposition 4**: *if Δ is a conditional knowledge base closed under CAND, OR, RW, CM, CUT, and CP, then the confidence relation it defines via (\*) is a partial acceptance order on disjoint events.*

**Theorem 1**: *The strict part of any confidence preorder that satisfies T, MI and Ac also satisfies CAND, RW, CM, CUT, OR, CP. If A > Ø whenever A ≠ Ø, the corresponding family of conditional assertions satisfies system P but for A = Ø.*

The main difference between Friedman's and Halpern's axiomatic and ours comes from a different understanding of system P on our side. In (Friedman and Halpern 1996), the need for A3 comes from the interpretation of φ ⊢ ψ as : φ ⊢ ψ ⇔ ([φ] ∩ [ψ] > [φ] ∩ [¬ψ] or [φ] ≡ Ø). Our simpler interpretation φ ⊢ ψ ⇔ ([φ] ∩ [ψ] > [φ] ∩ [¬ψ]) allows us to drop A3. Actually, we never assume that the inference φ ⊢ ψ holds when [φ] is empty or impossible and we drop the reflexivity φ ⊢ φ of system P.

## 4. ADDITIVE AND NON-ADDITIVE ACCEPTANCE PREORDERS

Friedman and Halpern (1996) show that, if ≥ is a transitive relation that satisfies MI, Ac and A3, the *strict* part of ≥ restricted to *disjoint* events can be represented by a family of complete acceptance preorders that also satisfy these properties. From Theorem 1, it is easy to show that it is still the case if even A3 is not assumed[4]. But this representation of an acceptance preorder by means of families of complete preorders proposed in (Friedman and Halpern 1996) is not fully satisfying; the comparison of non-disjoint events cannot be deduced from the comparison of disjoint events, unless the acceptance order is supposed to be additive. Recall that additivity means that, for any A, B and C such that A ∩ (B∪C) = Ø, A ∪ B ≥ A ∪ C ⇔ B ≥ C. If ≥ is additive, then, for any A, B, C such that A∩(B∪C) = Ø:

- A ∪ B ≡ A ∪ C ⇔ B ≡ C;
- A ∪ B > A ∪ C ⇔ B > C.

If additivity is assumed, the relation > can thus be soundly reconstructed as well as an acceptance preorder, using Proposition 1, but *this is not necessarily the original one*. In other terms, the acceptance preorders that can be soundly recomputed from the knowledge of > among disjoint events are those such that two disjoint events are never equivalent. Otherwise, the distinction between equivalence and incomparability can be lost.

However, in the context of acceptance, there is a strong incompatibility between the existence of equivalent disjoint events and additivity. Indeed, acceptance preorders have the following highly possibilistic property:

**Proposition 5**. *Let A, B and C be three disjoint events. If ≥ is an acceptance preorder, then:*
$$C \equiv A > B \Rightarrow A \equiv A \cup B \equiv C \equiv C \cup B > B$$

Proof: Since A > B and C ≡ A , then C ∪ A > B and C ∪ B ≥ A. It is impossible to have C ∪ B > A (otherwise, by Ac, C > A ∪ B and thus C > A). Then C ∪ B ≡ A, i.e. by transitivity C ∪ B ≡ C.

In other terms, if C ≡ A > B, then B is negligible when compared to A. Additivity rules out the possibility of handling negligibility since additivity implies that:
if B > Ø, then for any A disjoint from B, A∪B > A    (\*\*)

**Proposition 6**. *if ≥ is an acceptance preorder satisfying (\*\*), then for any three disjoint events A, B and C:*
$$C \equiv A > B \Rightarrow B \equiv \emptyset.$$
*Thus, if ≥ is additive, C ≡ A > B ⇒ B ≡ Ø.*

Proof : C ≡ A > B ⇒ A ≡ A ∪ B . By (\*\*), B ≡ Ø. Moreover, any additive relation satisfies (\*\*).

Hence as soon as equivalence is allowed between disjoint events, additive accepance orders have to be very particular: no state can be lower than two equivalent states, except if it is null. In summary, assuming additivity + acceptance rules out most of the numerical theories of uncertainty, that define a *complete* preorder between events, i.e. where disjoint events can be equivalent. Moreover, the notion of additivy is incompatible with the notion of negligibility that pertains to several theories of uncertainty (e.g., Lehmann, 1996) .

So, if we do not want to restrict our framework to additive confidence relations, we have to extend the representation theorem, showing that any (possibly partial) acceptance preorder is *equivalent* to a family of complete acceptance preorders. The fact that a partial preorder is equivalent to a family of preorders is obvious. The important point is that the semantics of acceptance partial preorders is defined in terms of complete *acceptance* preorders, be the original partial order additive or not.

**Theorem 2**: *Any acceptance preorder ≥ can be represented by a family F(≥) of complete acceptance preorders ≥$_f$ such that: A ≥ B ⇔ ∀ ≥$_f$ ∈ F(≥), A ≥$_f$ B.*

---

[4] Indeed, the restriction of > to disjoint events is equivalent to a consistent closed base of defaults, and such a belief base can be represented by a family of possibility measures (following Dubois, Prade (1995a)), *even if RR is not assumed*: if A ∪ B > A ≡ B ≡ Ø for some A and B, a possibility of 0 is assigned to A in some of the possibility measures, a possibility of 0 is assigned to B in the other ones : hence, it may hold that A ∪ B > Ø although A ≡ B ≡ Ø.

**Sketch of the proof:** First, we show that for any acceptance preorder ≥ there is a complete acceptance preorder ≥' such that A ≥ B ⇒ A ≥' B and A > B ⇒ A >' B. ≥' *agrees* with ≥, in the sense that it transforms some incomparable events A ⋈ B into comparable ones (in fact, A ⋈ B into either A > B or A < B). ≥' is said to be more discriminant than ≥ iff it agrees with it and ∃A,B such that A ⋈ B and A >' B. The idea is to introduce for some selected A ⋈ B pair a new constraint of the form A >'B and close the obtained more discriminating relation via transitivity and Ac. Since this closure provably does not introduce any contradiction (such as B >' A), the procedure continues by selecting a new C ⋈ D pair and so on, until no such pair remains. Considering the family **F**(≥) of complete preorders that the procedure can generate by changing the selection of A ⋈ B pairs and deciding B > A instead of A > B, we prove the theorem above.

The above result still holds is we restrict to the subfamily of maximally discriminant complete acceptance preorders. Conversely, it is easy to show that:

***Proposition 7:*** *Let F be a non empty set of complete acceptance preorders such that* $\forall \geq_f, \geq_g \in F, A \equiv_f B \Rightarrow A \equiv_g B$. *The order* $\geq_F$ *defined by:* $A \geq_F B$ *iff* $\forall \geq_f \in F, A \geq_f B$ *is an acceptance preorder.*

These results extend the ones of (Friedman and Halpern 1996), and generalizes the ones of Benferhat et al. (1997a) that capture system P by linear possibility orders.

## 5. NUMERICAL REPRESENTATIONS OF ACCEPTANCE

The previous result shows that the definition of a partial confidence order that satisfies deductive closure of accepted beliefs comes down to the definition of a set of *complete* acceptance preorders. Our generalisation is thus in accordance with usual semantics of system P, although more general. But this is of no use in terms of cost of representation, since representing many preorders can be as costly as representing a partial order. One of the main interests of usual confidence measures is that they can be defined more economically, from a distribution (a mass assignment for Shafer's belief functions, a possibility distribution for possibility measures, etc.). That is why we focus here on the charaterization of numerical confidence measures that agree with deductive closure.

Since S is supposed to be finite, any acceptance complete preorder can be mapped to a numerical scale. We recover *context tolerant acceptance functions* (Dubois, Prade 1995b):

***Definition 3:*** *An acceptance function f is a [0,1]-valued set function such that*
- $f(S) = 1$ *and* $f(\emptyset) = 0$;
- *if* $A \subseteq B$, $f(B) \geq_R f(A)$;
- *if* $f(A) >_R f(\bar{A})$ *and* $f(B) >_R f(\bar{B})$ *then* $f(A \cap B) >_R f(\bar{A} \cup \bar{B})$.

*An acceptance function is said to be context-tolerant iff*
- $f(A \cup B) >_R f(C)$ *and* $f(A \cup C) >_R f(B) \Rightarrow f(A) >_R f(B \cup C)$

*where* $>_R$ *and* $\geq_R$ *are the usual orders on real numbers.*

   *A context tolerant acceptance function f obviously defines a normalized acceptance complete preorder* $\geq_f$ *as follows:* $A \geq_f B$ *iff* $f(A) \geq_R f(B)$.

Well-known confidence relations are usually specialized by specifying a stability property on the relation A ≥ B with respect to adding new elements to sets A and B or deleting common elements: the additivity property comes from the conjunction of two weaker properties (Dubois, 1986):

***Definition 4.*** *Let A, B and C be such that* $A \cap (B \cup C) = \emptyset$.
- ≥ *is a relation of type OR iff:* $B \geq C \Rightarrow A \cup B \geq A \cup C$
- ≥ *is a relation of type AND iff:* $A \cup B \geq A \cup C \Rightarrow B \geq C$
- ≥ *is additive iff it is of both types AND and OR.*

Comparative probabilities are additive confidence relation. Possibility orders are of type OR, while the dual orders (necessity orders) are of type AND : they are generally not additive. Weakest versions of type AND and OR can be given, that are characteristic of Shafer's belief functions and plausibility measures (Wong et al 1991):

***Definition 5.*** *Let A, B and C be three disjoint events.* ≥ *is a relation of type*
Weak AND iff: $A \cup B > B \Rightarrow A \cup B \cup C > B \cup C$.
Weak OR iff: $A \cup B \cup C > B \cup C \Rightarrow A \cup B > B$

In fact, only few of these confidence measures are compatible with deductive closure without any restriction. As previously stated, necessity measures and possibility measures are fully in accordance with the axiomatics of acceptance. In contrast, only probability measures P such that $\forall A \exists s \in A$ such that $P(\{s\}) > P(A \setminus \{s\})$ are context-tolerant acceptance functions. This leads to very special probability measures generally inducing a linear order on states, and such that the probability of a state is much bigger than the probability of the next probable state (Snow 1994, Benferhat et al 1997a) - they can be called "big-stepped" probabilities. Namely, let $S = \{s_1,...,s_n\}$ with $P(\{s_1\}) > ...> P(\{s_{r-1}\}) \geq P(\{s_r\}) > P(\{s_{r+1}\}) = ... = P(\{s_n\}) = 0$, then P is a context-tolerant acceptance function iff: $\forall i < r - 1, P(\{s_i\}) > \sum_{j=i+1,n} P(\{s_j\})$.

This situation is related to property (**), that is actually induced by Weak AND property. Indeed, as soon as the Weak AND property is assumed, Proposition 6 holds. Thus $C \equiv A > B \equiv D > \emptyset$ cannot be expressed for mutually disjoint events using a set function that satisfies weak AND. In the context of a complete preorder, this property has drastic effects since it enforces a linear ordering on the non impossible states. In other terms, acceptance complete preorders of this type correspond to very special confidence functions.

This restriction not only applies to qualitative probabilities, that obviously satisfy Weak AND, but also to Shafer 's belief functions (the weak AND relation is actually characteristic of Shafer's belief functions). The class of belief functions obeying the acceptance axioms can be precisely specified in terms of the structure of the focal sets (sets with positive mass). First, let us recall the following result about acceptance belief (resp. plausibility) functions (indeed any context-tolerant acceptance belief (resp. plausibility) function is an acceptance belief (resp. plausibility) function).

**Proposition 8** (Dubois, Prade 1995b):

A belief function Bel (as well as its dual plausibility function Pl), with basic probability assignment m, is an acceptance function based on a kernel $K = \bigcap \{A, Bel(A) > Bel(\bar{A})\} = \bigcap \{A, Pl(A) > Pl(\bar{A})\}$ if and only if:

i) either $K$ is a singleton s.t. $m(K) > \sum_{A \subseteq \bar{K}} m(A)$ (then $K$ is the kernel and $|K| = 1$).

ii) or any focal subset $F$ of Bel is such that $F \supseteq K$ where $K$ is a focal subset such that $|K| \geq 2$

iii) or the only focal subsets are $\{\omega_K\}$, $\{\omega'_K\}$ with $m(\{\omega_K\}) = m(\{\omega'_K\})$, and possibly supersets of $K = \{\omega_K, \omega'_K\}$.

Let us call *minimal focal set* a focal set that does not contain other focal sets. It can be shown that:

**Proposition 9:** A belief function defines an acceptance preorder (i.e., is a context tolerant acceptance function) iff the following requirements hold:

• Only one minimal focal set can contain more that one element (other ones must be singletons).

• The minimal focal sets are strictly ordered and the less important one is the one that is not a singleton. In other terms, the mass assignment for minimal focal sets is of the form: $m(\{s1\}) > .... > m(\{si\}) > m(\{si+1, ..., sk\}) > m(\{sk+1\}) = ... = m(\{sn\}) = 0$ where $S = \{s1, ..., sn\}$ is the set of states and the sum of masses is one.

• The mass assignment defines a "big-stepped" belief function, i.e., for any minimal focal set $\{sj\}$: $m(\{sj\}) = Bel(\{sj\}) > Bel(\{sj+1, ..., sm\}) = \sum_{A \subseteq \{sj+1,..,sm\}} m(A)$. Necessity measures are among this class: they have a single minimal focal set. But some acceptance-like belief functions are not necessity measures.

Hence, belief functions are rather incompatible with the requirement of acceptance. But the dual measures - plausibility measure ou equivalently confidence preorders of type weak OR are not constrained by the trivializing Proposition 6. The study of these context-tolerant acceptance plausibility functions cannot derive from the study of the belief functions, because the dual of a *context tolerant* acceptance function is not necessarily a *context tolerant* acceptance function (although the dual of an acceptance function is an acceptance function).

The above Proposition 9 pertains actually to the geometric conditioning of belief functions (i.e., $Bel(B|A) = Bel(A \cap B)/Bel(A)$), since the requirement of context-tolerant acceptance is applied to the belief relation. Applied to a plausibility relation, the requirement exploits Dempster rule of conditioning (i.e., $Pl(B|A) = Pl(A \cap B)/Pl(A)$). To obtain context-tolerant acceptance plausibility functions, it is enough to start from Proposition 8 which specifies all plausibility functions which are acceptance functions, and then to require that the Dempster conditioning of such functions should yield an acceptance function again. This method provides us with generic examples of context-tolerant plausibility functions:

**Example 1:** There are $k \geq 0$ minimal focal elements which are singletons. After a suitable renumbering of the elements of S assume $m(\{s_1\}) > ...>m(\{s_k\})$, and the other focal elements $E_h$ form a nested structure. Moreover the following condition should hold:

$\forall i, m(\{s_i\}) > \sum_{j=i+1,k} m(\{s_j\}) + \sum_{E_l} m(E_l)$,

The $E_l$ in the sum are those $E_h$ which do not contain elements $s_1$ to $s_i$.

**Example 2:** The structure of the focal elements is centered in the following way: there is a kernel focal element K and the other focal elements are either nested inside K, or contains K and are of the form $F= K \cup \{s_i\}$ where the $s_i$'s are distinct elements of S which are not focal elements; moreover they should satisfy the condition:

$\forall i, m(K \cup \{s_i\}) > \sum_{j=i+1,r} m(K \cup \{s_j\})$, with $m(K \cup \{s_1\}) > ...>m(K \cup \{s_r\})$.

These two examples illustrate the fact that context-tolerant plausibility functions combine big-stepped probabilities and possibility-like nested structures. It is also worth pointing out that extremely simple basic mass assignment structures do not yield context tolerant acceptance plausibility functions. For instance, consider a structure made of only two disjoint focal elements. Then at least one should be a singleton and have a mass strictly greater than the other (except if the other is itself a singleton), in order to define a context tolerant plausibility.

# 6. SUMMARY AND CONCLUSIONS

The most general, yet meaningful, representations of partial belief are preorders of events monotonic with inclusion, that can be viewed as families of complete preorders, hence as families of confidence functions. The only useful part of a preorder on events for the purpose of capturing the set of accepted beliefs is its strict part restricted to disjoint events, and requesting that it be deductively closed enforces the Ac axiom, which in turn enforces all properties of nonmonotonic preferential inference. We thus agree with Friedman and Halpern (1996) that give a central role to the CAND axiom: in the context of very natural requirements on the order on events (T and MI), the axiom of deductive closure enforces the other properties of system P (except RR, that is not required here).

This leaves little room for most numerical theories of uncertainty. Most probability functions, most plausibility functions and most belief functions fall out of the range of acceptance functions. This contrasts with possibility theory that is fully consistent with a logical handling of accepted beliefs.

However, there exist acceptance structures which are more general than possibility measures, not fully characterized yet, which are complete acceptance orders. Indeed, as explained in Section 4 (see Proposition 5), acceptance orders actually obey a highly "possibilistic" property: for any three disjoint events A, B and C, if $\geq$ is a complete acceptance preorder, then $C \equiv A > B \Rightarrow A \equiv A \cup B \equiv C \equiv$

C ∪ B > B. When considering *complete* preorders, like those induced by numerical representations, it can hardly be assumed that no pair of disjoint events (e.g. no pair of states) is equivalent. Thus there may be some non null events such that A ≡ A∪B. In other terms, either the ordering never allows to have equivalent disjoint events, and is thus restricted to very to very special confidence functions, or it assumes that some events are negligible in front of others (here B, in front of A and C).

These results are quite negative regarding the general compatibility between logical representation of beliefs, viewed as accepted propositions, that use the usual notion of deductive closure for representing implicit beliefs, and numerical representations based on set-functions for reasoning about partial belief. It is a severe impediment to a generalized view of theory revision based on orders on formulae derived from uncertainty theories other than epistemic entrenchments, simply because the assumption of closed belief sets is devastating in this respect.

One possible conclusion is that the notion of closed belief set is not fully adapted to the modelling of belief, not only because this notion disregards syntactic aspects (as advocated by tenants of syntactic revision and opponents to the logical omniscience assumption), but because the closure under conjunction is not always intuitively plausible when reasoning with partial beliefs (as already revealed in the lottery paradox). Weaker types of "deductive closures" might be considered for this purpose as for instance unions of standard deductively closed sets of propositions (that may be globally inconsistent). This type of closure is encountered in argument-based reasoning under inconsistency (Benferhat et al 1997b). Tolerating inconsistency is indeed incompatible with standard deductive closure. It turns out that most confidence measures (and noticeably probabilities) synthetize partially conflicting pieces of information while possibility measures do not. It may explain why the latter seem to be the only simple ones that sanction the concept of belief set, where inconsistency is banished.

# References

S. Benferhat, D. Dubois, H. Prade (1997a). Possibilistic and standard probabilistic semantics of conditional knowledge. *Proc. 14th National Conf. on Artificial Intelligence* (AAAI-97), Providence RI, 70-75, 1997.

S. Benferhat, D. Dubois, H. Prade (1997b). Some syntactic approaches to the handling of inconsistent knowledge bases: a comparative study. *Studia Logica* 58: 17-45.

R.T. Cox (1961) *The algebra of probable inference*. The John Hopkins Press, Baltimore.

D. Dubois (1986) Belief structures, possibility theory and decomposable confidence measures on finite sets. *Computers and Artificial Intelligence*, 5(5), (Bratislava), 403-416.

D. Dubois, H. Fargier, H. Prade (1997). Decision making under ordinal preferences and comparative uncertainty. *Proc. of the 13h Conf. on Uncertainty in Articicial Intelligence*, Providence, August 1997, 157-164.

D. Dubois, H. Prade (1992). Belief change and possibility theory. In *Belief Revision*, P. Gärdenfors, ed., Cambridge university press, 142-182.

D. Dubois, H. Prade (1995a). Conditional objects, possibility theory and default rules. In *Conditional: From Philosophy to Computer Science* (G. Crocco, L. Farinas del Cerro, A. Herzig eds.), Oxford University Press, 311-346.

D. Dubois, H. Prade (1995b), Numerical representation of acceptance. *Proc. of the 11th Conf. on Uncertainty in Articicial Intelligence*, Montréal, August 1995, 149-156

P. Fishburn (1986). The axioms of subjective probabilities. *Statistical Science*, 1, 335-358.

N. Friedman, J. Halpern (1996) Plausibility measures and default reasoning. *Proc of the 13th National Conf. on Artificial Intelligence* (AAAI'96), Portland, OR, 1297-1304. To appear in *J. Assoc. for Comp. Mach.*

P. Gärdenfors (1988) *Knowledge in flux*. MIT press, Cambridge, MA.

J. Halpern (1996) Defining relative likelihood in partially-ordered structures *Proc of the 12th Conf. on Uncertainty in Artificial Intelligence* (UAI'96), 299-306.

K. Kraus, D. Lehmann, M. Magidor (1990) Nonmonotonic reasoning, preferential models and cumulative logics. *Artificial Intelligence*, 44, 167-207.

H. E. Kyburg. *Probability and the logic of rational Belief*. Wesleyan University Press. Middletown, Ct. 1961

H. E. Kyburg (1988) Knowledge. *Uncertainty in artificial intelligence* vol 2, J. F. Lemmer and L. N. Kanal eds, Elsevier, 263-272.

D. Lehmann (1996) Generalized qualitative probability: Savage revisited. *Proc. of the 12th Conf. on Uncertainty in Articiciai Intelligence*, Portland August 1996, 381-388.

D. L. Lewis (1973) *Counterfactuals*. Basil Blackwell, Oxford, UK.

Shafer G. (1976) *A Mathematical Theory of Evidence*. Princeton University Press, Princeton.

P. Snow (1994) The emergence of ordered belief from initial ignorance. *Proc of the 12 th National Conference on Artificial Intelligence* (AAAI 94), 281-286.

S. K. M. Wong, Y. Y. Yao, P. Bollmann, H. C. Burger (1991) Axiomatization of qualitaitve belief structure. *IEEE Transactions on SMC*, vol 21 no 4, 726-734.

L. A. Zadeh (1978) Fuzzy sets as a basis for a possibility theory, *Fuzzy Sets and Systems*, 1, 3-28.